# Fixed-length Bit-string Representation of Fingerprint by Normalized Local Structures

Jun Beom Kho, Andrew B. J. Teoh, Wonjune Lee and Jaihie Kim

**Abstract**—In this paper, we propose a method to represent a fingerprint image by an ordered, fixed-length bit-string providing improved accuracy performance, faster matching time and compressibility. First, we devise a novel minutia-based local structure modeled by a mixture of 2D elliptical Gaussian functions in the pixel space. Each local structure is mapped to the Euclidean space by normalizing the local structure with the number of minutiae that associates to it. This simple yet crucial crux enables fast dissimilarity computation of two local structures with Euclidean distance without distortion. A complementary texture-based local structure to the minutia-based local structure is also introduced whereby both can be compressed via principal component analysis and fused easily in the Euclidean space. The fused local structure is then converted to a K-bit ordered string via a K-means clustering algorithm. This chain of computation with sole use of Euclidean distance is vital for speedy and discriminative bit-string conversion. The accuracy can be further improved by a finger-specific bit-training algorithm in which two criteria are leveraged to select useful bit positions for matching. Experiments are performed on Fingerprint Verification Competition (FVC) databases for comparison with existing techniques to show the superiority of the proposed method.

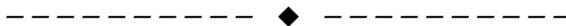

## 1 INTRODUCTION

CONVENTIONALLY a fingerprint template incorporates four types of fingerprint minutia features: the x- and y-coordinates, ridge orientation and the minutia type (termination or bifurcation). Fingerprint recognition using the minutia template is widely adopted in numerous applications compared to other biometric modalities because of its high distinctiveness, collectability, and low cost [1], [2]. For minutia-based fingerprint systems, the fingerprint image is only partially captured by a fingerprint scanner; therefore, the extracted minutiae are unordered, vary in size, and only their relative positions are known. Thus, conventional fingerprint recognition methods mostly rely on relative geometric minutia information, and to find the best minutia matching pairs [3] between two images, all possible combinatorial pairs should be considered.

Recently, the emergence of ubiquitous mobile computing, body sensor networks, the Internet of Things (IoT), financial technology, etc., has generated stringent expectations, which lead inapplicability of the conventional fingerprint recognition to current industrial needs. The most prominent difficulty in fingerprint recognition system is lacking of security [4]. For instance, as smartphones and IoT-enabled devices are always connected to network, the system is easily exposed to diverse attacks [5], which cause permanent loss of private biometric information. According to [6], [7], [8], [9], original fingerprint images could be reconstructed from their geometrical minutia data, so if a hostile entity compromises the fingerprint template, the biometric identity cannot be used permanently [10]. For secure protection of fingerprint templates, a cancelable fingerprint template [11], [12], [13], [14] and bio-encryption [15], [16] have been proposed [17], [18], however, some of these solutions can only be adopted when the fingerprint is represented by an ordered and fixed-length bit-vector [13], [14], [19]. In addition, these devices have physical limitations, such as low computing power and small sensor size, so faster and better performance is required. However, this may not be achievable using a conventional minutia-based fingerprint recognition system because of its complex matching algorithm. Therefore, re-examination of conventional fingerprint recognition based on minutia representation is required.

As an alternative approach to overcome the limitations of minutia-based matching algorithms, minutia data have been converted into binary bit-strings, although their bit-string representations remain unordered and variable in size in the approach. The minutiae cylinder code (MCC) [20] is a well-known example, with a cylinder-shaped space being constructed to represent the relative positions and directions of a central minutia and neighbors. The MCC exhibits decent recognition accuracy after representation of each minutia by a binary bit-string. However, a complex matching mechanism is needed to find the best-matching minutia pairs. In addition, as the number of bit-strings depends on the number of extracted minutiae, a considerably large storage space is required. Abe and Shinzaki [21] have proposed a minutiae relation code (MRC) to express a fingerprint template as a set of vectors representing relative differences in the positions and directions of minutia pairs, reporting superior accuracy to

———————————————
- J. Ko, A. Teoh, and J. Kim are with School of Electrical and Electronic Engineering, Yonsei University, Seoul 120-749, Korea.
  E-mail: kojb87@hanmail.net, bjteoh@yonsei.ac.kr, jhkim@yonsei.ac.kr
- W. Lee is with Hyundai Mobis, 17-2, Mabuk-ro, 240 beon-gil, Giheung-gu, Geyonggi, 16891, Korea. E-mail: wj.michael@mobis.co.kr

Corresponding author: Jaihie Kim

MCC. However, MRC binary representation is also unordered and variable in size, having similar problems to the MCC. Several other methods to convert minutia data into binary bit-strings have been proposed [11], [15], [22], [23], [24], [25], [26], [27], but the difficulties with complex matching and template protection scheme application remain unresolved.

## 1.1 Related Work

Previous studies on ordered and fixed-length bit-strings for fingerprint representation can be classified into two categories [28]: alignment-based and alignment-free approaches. The former aligns fingerprint images based on a global reference such as the fingerprint core point, yielding an ordered and fixed-length bit-string based on that point. Previously, Luo et al. [28] proposed template point acquisition by combining existing singular and focal point extraction methods, aligning fingerprint images based on the template point. Then, they converted a fingerprint image into a fixed-length bit-string via a global minutia cylinder code.

Further, Jain et al. [29] proposed the FingerCode, with fingerprint images aligned by core points and segmented into 80 sectors. Eight directional Gabor filters were then applied to obtain the fixed-length Gabor response vectors. Nandakumar [30] also aligned fingerprint images using fingerprint focal points, and applied the Fourier transform to represent minutia features as the binarized phase spectrum. The main difficulty for all these alignment-based methods is misalignment due to the global reference location error. The core or focal points may also be unavailable in fingerprint images, as undecided partial fingerprint images are captured by conventional touch-based fingerprint scanners.

The alignment-free approach utilizes relative positional information among adjacent minutiae to obtain the local-feature frequency histogram, or quantizes the feature values obtained from correlation of adjacent minutiae for fixed-length bit-conversion of a fingerprint image. Bringer and Despiegel [31] proposed a binary bit-conversion method using the minutia vicinity, which was constructed using the positional relation between a central minutia and its neighboring minutiae within a predefined radius. From a large training fingerprint database, a subset of representative vicinities was predefined, and the vicinities of the query fingerprint were compared to the overall representative vicinities to find the best match. Then, the corresponding bit position was set to "1." However, the exponential computation between each of the input minutiae and all the stored representative vicinities (50,000) entailed very high computation cost, and the recognition accuracy degraded if the input minutia could not find a close vicinity minutia.

As another example of the alignment-free approach, Wong et al. [32] proposed a multi-scale bag-of-words system for fixed-length bit-conversion of a fingerprint image. The minutia local descriptor was formulated by multi-line code, and representative local descriptors were selected by K-means clustering in a fine-to-coarse manner. Then, a dynamic quantization technique was employed to allow more bits to be assigned to the more discriminative feature components. However, it may not be possible for this algorithm to be generalized to fingerprints outside the training set. Further, Xu and Veldhuis [33] proposed a spectral minutia approach to convert an unordered minutia set into a fixed-length bit-string. They applied a Fourier transform to a minutia set and remapped the Fourier spectral magnitudes onto the polar-logarithmic coordinate. Then, the extracted feature values were converted into a fixed-length bit-string by quantization. However, the recognition accuracy was found to be severely degraded by the quantization error. Farooq et al. [34] represented local minutia sets as triangular features composed of seven factors (the lengths of three sides, the three angles between each side and each minutia orientation, and the triangle height). Then, each triangular feature value was quantized into 24 bits, and each fingerprint image was expressed as $2^{24}$ bits. However, this method had high computational cost due to the exhaustive calculation of all triangular sets obtained from the fingerprint image.

Jin et al. [4] utilized kernel principal component analysis (KPCA) and a heat kernel function to construct the KPCA

TABLE 1
SUMMARY OF THE FINGERPRINT BIT-STRING CONVERSION METHODS

| Category | | Method | Type | Restriction |
|---|---|---|---|---|
| Unordered, variable-sized bit-string conversion | | [11],[15],[20], [21],[22],[23], [24],[25],[26], [27] | Binary | - Requires combinational matching<br>- High computational cost<br>- Large memory storage<br>- Improper to certain biometric cryptographic applications |
| Ordered, fixed-length bit-string conversion | Alignment-based approach | [28],[29] | Binary | - Requires pre-alignment<br>- Vulnerable to image rotation and translation<br>- Low recognition accuracy compared with conventional minutia matching |
| | Alignment-free approach | [4],[14],[31], [33],[34],[35], [41] | Binary | - Low recognition accuracy compared with conventional minutia matching<br>- Difficult to implement |

3projection matrix based on matching scores of fingerprint pairs from a training dataset. Then, dynamic quantization was adopted to allocate binary bits to each feature component according to the user-specific feature discriminability. The experimental results showed superior performance to existing methods, but the scheme could not be generalized beyond the fingerprint images captured in the training data. Finally, Wang et al. [14], [35] converted a fingerprint image into a fixed-length bit-string by quantizing the relative difference of the distance and orientation among randomly selected minutia pairs and assigning one bit to each quantized value. This method produces a fixed-length bit-string easily, but the discriminative power of the minutia feature can be degraded severely during the quantization process. Large storage space is also required. A summary of the existing methods for fixed-length bit-string conversion is provided in Table 1.

### 1.2 Motivations and Contributions

As reasoned above, a new type of fingerprint representation is urgently required to accommodate contemporary fingerprint recognition applications, which impose higher accuracy, speed, storage, and security requirements. As discussed in Section 1.1, the alignment-free approach to fingerprint bit-string representation has been widely explored, as it is less prone to image rotation and translation than the alignment-based approach and generally demonstrates better recognition accuracy. However, several issues require resolution:

1. Simple similarity/dissimilarity measure of minutia pair: In the previous methods [20], [31], to compute similarity/dissimilarity score of two central minutiae, a complicated geometric computation is required to compare all neighbor minutia pair located in the local structures. Such a complex computation makes the clustering method difficult to be applied for the fingerprint bit-conversion.
2. Compensation for recognition accuracy performance degradation: Generally, information loss occurs upon conversion of a fingerprint image to a fixed-length bit-string; this is caused by loss of minutia information, quantization errors, defective grouping of similar local descriptors, matching errors between predefined groups and query minutiae, and unused global minutia information.
3. Small-sized bits for low storage requirement with simple and fast matching: Existing fixed-length bit-string conversion methods produce excessively long bit-strings (e.g., 50,000 bits in [31], $2^{24}$ bits in [34]), through complex conversion processes. Therefore, they require large fingerprint template storage space.
4. Generalization beyond training set: Alignment-free methods can be training-free [13], [14], [19], [31] or training-based [4], [32]. The latter generally outperforms the former in terms of accuracy. Unfortunately, the existing training-based methods cannot generalize beyond the training set.

In this paper, we propose a fixed-length and ordered bit-string conversion method to overcome the above difficulties. This study makes the following contributions:

1. Due to the proposed normalized local structure, two central minutiae in the two local structures are now resided in the Euclidean space, and their dissimilarity can be computed by the Euclidean distance (ED) in pixel-based space. In addition, ED of minutia pair represented by normalized local structures follows the fingerprint recognition convention that rewards matched minutia pair more than the penalty of unmatched minutia pair.
2. In our work, each minutia (reference minutia) in a fingerprint image is represented by a collection of neighboring minutiae enclosed within a circle. Each neighboring minutia is modeled by a two-dimensional elliptical Gaussian function (2DGF). The mixture of multiple 2DGF in the local circular region is then normalized by the number of neighboring minutiae. We refer to this structure as a *minutia-based local structure (MBLS)*. The MBLS reflects the difficulty in estimating minutia positions due to skin elastic distortion and minutia extraction errors depending on distance from the reference minutiae.
3. A texture-based local structure (TBLS) is devised and fused with the MBLS via a weighted feature-level fusion scheme to improve the intermediate fingerprint representation discriminative before bit-string conversion. To our knowledge, texture-based information was not used for fingerprint bit-string conversion in previous works.
4. A K-means algorithm is applied to transform the fused MBLSs and TBLSs into a fixed-length bit-string. The conversion scheme does not suffer from the poor generalization that hinders existing methods. The bit-string is also compressible without significant performance degradation, because of its sparse representation nature.
5. A finger-specific bit-training mechanism is proposed to further enhance the accuracy on fingerprint bit-strings. Two training criteria, i.e., discrimination power and reliability, which respectively capture the inter-class and intra-class variances of the fingerprint bit-string, are leveraged to select useful bit positions for bit-string matching.

In a nutshell, the *normalized* mixture of Gaussian representation of minutiae and texture local structures that reside in the Euclidean space, have greatly facilitated the feature-level fusion, PCA subspace reduction and clustering. This chain of processing is crucial for speedy and high discriminative bit-string conversion.

The remainder of this paper is organized as follows: Section 2 describes the proposed MBLSs and TBLSs, the fixed-length bit-conversion, and the finger-specific bit-training in detail. Experimental results with analysis are presented in Section 3, and a conclusion and discussion of future works are given in Section 4.

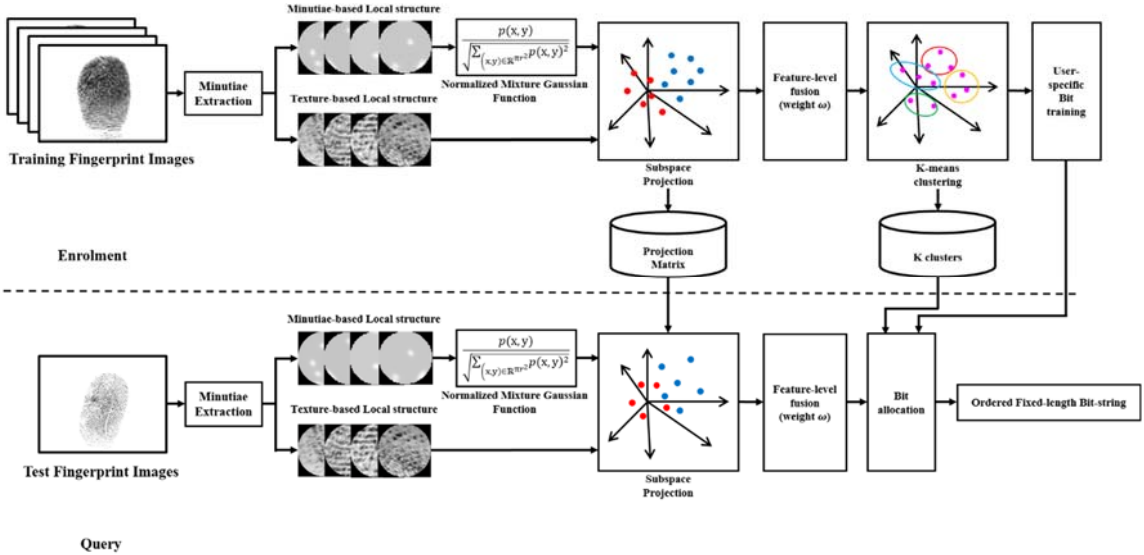

Fig. 1. Overall flows of the proposed method

## 2 PROPOSED METHOD FOR ORDERED AND FIXED-LENGTH BIT-STRING CONVERSION

### 2.1 Overview of Proposed Method

In the proposed method (PM), the minutiae of a fingerprint image are first extracted. For each minutia set as the reference minutia, each neighboring minutia within a circle of radius $r_m$ centered at the reference minutia is modeled by the 2DGF. The mixture of all 2DGFs is normalized by the total number of neighbor minutiae enclosed by the local circular region; we call this the MBLS. An MBLS can be expressed in vector form, with its size being equal to the pixel number in the circular region. The dissimilarity of two local structures is measured by the ED. Through the ED, subspace reduction and clustering of similar minutiae become possible, allowing simplification of the dissimilarity computation as well as dimensionality reduction, with low computation cost and bit requirements.

In addition to the MBLS, the texture information around the minutia is expressed by the TBLS, of which the circular boundary is smaller than that of the MBLS; the dissimilarity is also computed via the ED. Subspace projection through principle component analysis (PCA) is applied to both local structures for dimensional reduction. Then, the two structures are combined via weighted feature-level fusion.

Numerous local structures derived from training datasets are assembled into $K$ groups by the K-means clustering algorithm with the ED metric. To represent a fingerprint image by a $K$-bit ordered string, the $d$-th bit in the bit-string is set to 1 if any minutia in the fingerprint image is captured by the $d$-th cluster, where $d = 1,…,K$. The accuracy can be further improved by the finger-specific bit-training algorithm, where multiple enrolled images of a finger are trained to utilize the discriminative power and reliability of each bit position separately. Fig. 1 shows the flow of the PM.

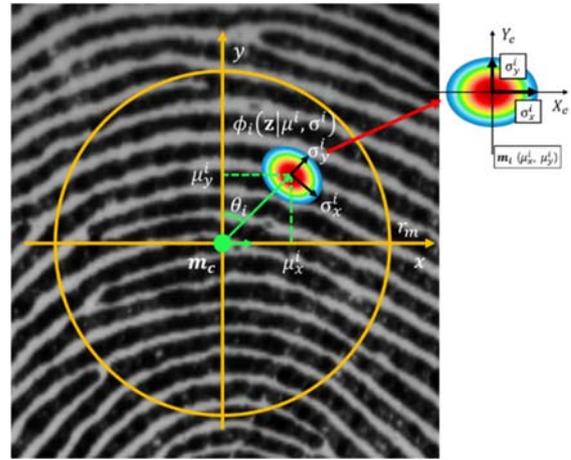

Fig. 2. 2D Elliptical Gaussian function of minutia $m_i$

### 2.2 Minutiae-based Local Structure

In this subsection, we present the formulation of the MBLS. The two key components of this local structure, i.e., Gaussian function modeling for the neighboring minutiae of a reference minutia and its normalization mechanism, are detailed.

#### 2.2.1 Modeling of Neighbor Minutia

Given a minutia set extracted from a fingerprint image, a circle with $r_m$ pixels based on a chosen reference minutia $m_c$ as the center can be formed, as depicted in Fig. 2. Neighbor minutiae surrounding $m_c$ and enclosed by the circle can be modeled individually by the 2DGF. Suppose $m_i$ is a neighbor minutia and the 2DGF of $m_i$, i.e., $\phi_i(\mathbf{z}|\mu^i, \sigma^i)$ is expressed as follows in the ($x$, $y$) Cartesian pixel space, where $\mathbf{z} \ni (x,y)$:

$$\phi_i(\mathbf{z}|\mu^i, \sigma^i) = \exp\left(-\left(a(\mathrm{x} - \mu_x^i)^2 + 2b(\mathrm{x} - \mu_x^i)(\mathrm{y} - \mu_y^i) + c(\mathrm{y} - \mu_y^i)^2\right)\right), \quad (1)$$

where

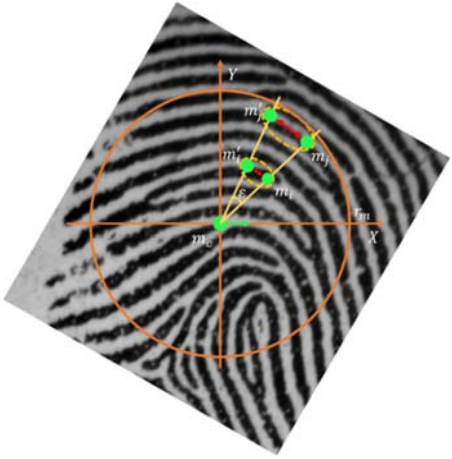

Fig. 3. Location errors of neighbors in local structure. When the direction of reference minutia $m_c$ is varied by $\varepsilon$, the position of minutia ($m_j$) is affected more strongly than that of minutia position ($m_i$).

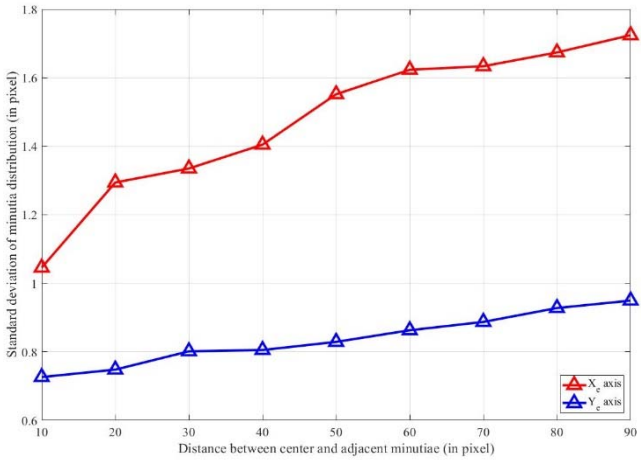

Fig. 4. Distributions of same minutiae in different images vs. distance from central minutia

$a = \frac{\cos^2 \theta_i}{2(\sigma_x^i)^2} + \frac{\sin^2 \theta_i}{2(\sigma_y^i)^2}, b = -\frac{\sin 2\theta_i}{4(\sigma_x^i)^2} + \frac{\sin 2\theta_i}{4(\sigma_y^i)^2}, c = \frac{\sin^2 \theta_i}{2(\sigma_x^i)^2} + \frac{\cos^2 \theta_i}{2(\sigma_y^i)^2}.$

Note that $\mu^i \ni (\mu_x^i, \mu_y^i)$ is the location of $m_i$ and $\sigma^i \ni (\sigma_x^i, \sigma_y^i)$ is the parameter to determine the $xy$ spreads of the 2DGF. The $\phi_i(\mathbf{z}|\mu^i, \sigma^i)$ of $m_i$ is aligned by the orientation of reference minutia $m_c$, i.e., $\theta_i$, and $X_e(Y_e)$ denote the direction of $\sigma_x^i$ ($\sigma_y^i$). To simplify the computation, $\phi_i(\mathbf{z}|\mu^i, \sigma^i)$ can be expressed in vector form with size $n_m = \lfloor \pi r_m^2 \rfloor$, as it is constrained in a circle of radius $r_m$. In this paper, we downscaled the local structure area to reduce the computational complexity.

$\phi_i(\mathbf{z}|\mu^i, \sigma^i)$ of each minutia has different $\sigma_x^i$ and $\sigma_y^i$ to reflect the $m_i$ positioning error, depending on the distance from $m_c$. As shown in Fig. 3, when a minutia is positioned further from the center, the minutia position can vary within the area extended to the horizontal axis (dotted line in Fig. 3). This is attributed to the alignment error of the local structure that rotated based on the direction of central minutia $m_c$. Thus, as shown in Fig. 3, minutia located further from the center will be affected more despite same amount of alignment error.

This intuition is verified by the experimental results depicted in Fig. 4 where $X_e$ and $Y_e$ indicate the directions of $\sigma_x^i$ and $\sigma_y^i$ in 2DGF of minutia $m_i$ (Fig. 2), respectively. When the distance between $m_i$ and $m_c$ increases, $\sigma^i$ becomes larger; this tendency is more apparent on the $X_e$ than the $Y_e$ axis. In our work, $\sigma^i \ni (\sigma_x^i, \sigma_y^i)$ is obtained experimentally.

### 2.2.2 Normalized Mixture 2D Gaussian Functions

For a minutia collection composed of $n$ minutiae enclosed by a circle of radius $r_m$ and centered at $m_c$, the minutiae can be collectively expressed by summing the $n$ 2DGF to obtain $p(\mathbf{z})$, which corresponds to the MBLS. Mathematically, the MBLS is defined as

$$p(\mathbf{z}) = \sum_{i=1}^{n} \phi_i(\mathbf{z}|\mu^i, \sigma^i) \in \mathbb{R}^{n_m}. \quad (2)$$

As $p(\mathbf{z})$ is expressed as a vector with size $n_m$, the dis-

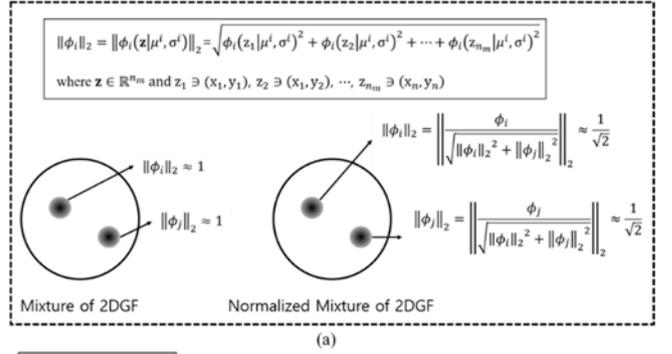

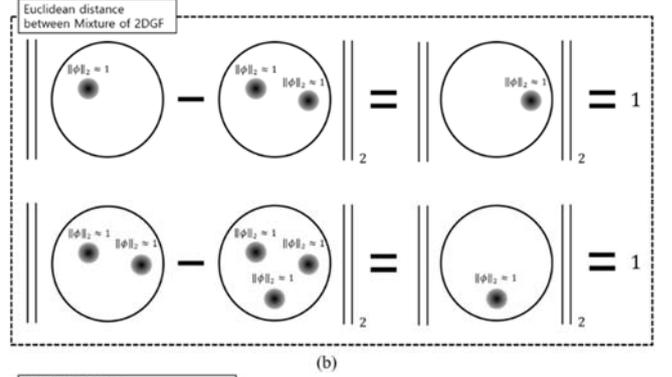

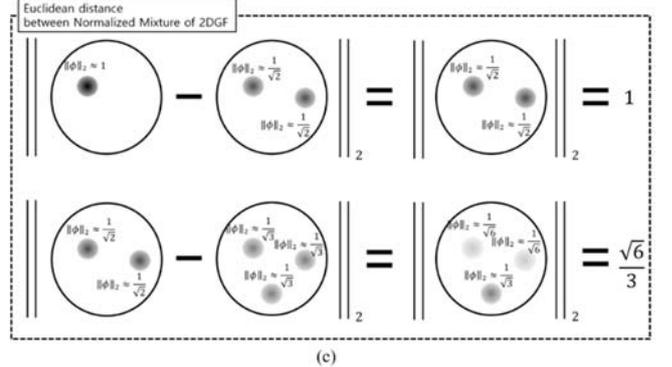

Fig. 5. An illustration of comparisons of Euclidean distances between minutia-based local structure pairs. (a) Assuming the scale of the area under a local structure is "1" ($\|\phi_i\|_2 \approx 1$ on left side), the scale of each normalized mixture of 2DGF depends on the number of minutiae contained in a local structure (depicted on right side). Although the lower pair is more similar in (b), the ED for the mixture of 2DGF is the same as that for the upper pair. The normalized mixture of 2DGF produces a small ED value for the lower pair in (c).

6similarity of a pair of $p(\mathbf{z})$ vectors can be measured conveniently using the ED. However, this tactic does not follow the fingerprint recognition convention that the award for matched minutia pairs has greater weight than the penalty for unmatched pairs. As illustrated in Fig. 5b, the ED of two instances of local structure matches are identical, but the lower instance should be recognized as the more similar match as there are more matched minutia pairs. To resolve this problem, $p(\mathbf{z})$ is normalized by the scale of the local structure; hence, an MBLS can be defined as follows:

$$p^N(\mathbf{z}) = \frac{p(\mathbf{z})}{\sqrt{\sum_{\mathbf{z}\in\mathbb{R}^{n_m}} p(\mathbf{z})^2}}. \qquad (3)$$

Thus, the dissimilarity measure of an MBLS pair can be modified to

$$\mathcal{S}(p_1^N, p_2^N) = \sqrt{\sum\sum_{\mathbf{z}\in\mathbb{R}^{n_m}} (p_1^N(\mathbf{z}) - p_2^N(\mathbf{z}))^2}. \qquad (4)$$

As a result, a small value of $\mathcal{S}(p_1^N, p_2^N)$ indicates that $p_1^N$ and $p_2^N$ have similar minutia distributions, and the center minutiae of two local structures are similar. As depicted in Fig 5c, the dissimilarity measure of a normalized mixture of 2DGFs by means of the ED coincides with the general convention that the award for a matched pair is much larger than the penalty for a non-matched pair or minutiae.

In addition, because the ED is employed, PCA subspace reduction can be used to reduce the feature dimensions of the MBLS. The ED is retained for the dissimilarity measure. This is because the EDs of the PCA-reduced dimension vector pairs and normalized mixture of 2DGF pairs are indeed equivalent:

$$\mathbf{z}_i = \mathbf{A}_M^T \mathbf{p}_i^N,$$

$$\begin{aligned}d_e(\mathbf{z}_i, \mathbf{z}_j) \\ = \sqrt{\left(\mathbf{A}_M^T\left((\mathbf{p}_i^N - \mathbf{m}) - (\mathbf{p}_j^N - \mathbf{m})\right)\right)' \mathbf{A}_M^T\left((\mathbf{p}_i^N - \mathbf{m}) - (\mathbf{p}_j^N - \mathbf{m})\right)} \\ = \sqrt{(\mathbf{p}_i^N - \mathbf{p}_j^N)'(\mathbf{p}_i^N - \mathbf{p}_j^N)} = d_e(\mathbf{p}_i^N, \mathbf{p}_j^N),\end{aligned} \qquad (5)$$

where $\mathbf{A}_M^T$ is a PCA projection matrix and $\mathbf{m}$ is an average vector of the normalized 2DGF mixture.

## 2.3 Texture-based Local Structure

In addition to the MBLS, texture information surrounding a minutia is equally beneficial for fingerprint matching. To our knowledge, texture information has not previously been considered for the ordered and fixed-length bit-string conversion of a fingerprint. In this paper, we use raw fingerprint image texture information directly, as we simply wish to demonstrate the feasibility of combination with the MBLS. However, well-established texture descriptors such as the histogram of gradient and local binary pattern [3], [29] can be utilized.

### 2.3.1 Pre-processing of Fingerprint Images

The TBLS utilizes the pixel intensity information within a given radius (different from the radius of the MBLS) from a chosen central minutia. However, as the pixel intensity can be badly affected by the image-acquisition environment, such as the sensor type or individual skin characteristics, captured fingerprint images are normalized before production of the TBLSs. Thus, in our work, a simple image normalization is applied to provide images with the same mean and standard deviation in the intensity values, before pixel information processing.

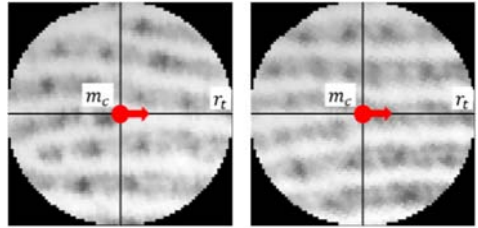

Fig. 6. Examples of the texture-based local structure

### 2.3.2 Generation of Texture-based Local Structure

As shown in Fig. 6, a circular local structure surrounding each minutia with radius $r_t$ is first taken, and the local structure is aligned to the origin by the direction of the central minutia $m_c$. The radius $r_t$ of the TBLS must be smaller than that of the MBLS. This is because the pixel intensity feature is weak at local structure misalignment, especially far from $m_c$. However, the TBLS captures discriminative feature details even within a small area. Therefore, selection of the appropriate radius for the TBLS is important, and the optimal radius used in this paper was obtained experimentally.

When the radius of the TBLS is set to $r_t$ pixels, the structure can be considered as a point in the $n_t = \lfloor \pi r_t^2 \rfloor$ dimensional space. Then, measuring the dissimilarity between two TBLSs via the ED is feasible, as for the MBLSs. However, the local texture size, $n_t$ = 5,000 generates excessive computational cost; thus, PCA-based subspace projection is applied to reduce the feature space. The dissimilarity between the two local structures in the PCA space is still measured by the ED. As shown in Fig. 7, top eigenvectors represent the texture features with reduced image noise; these are combined with the MBLS for fingerprint bit-conversion.

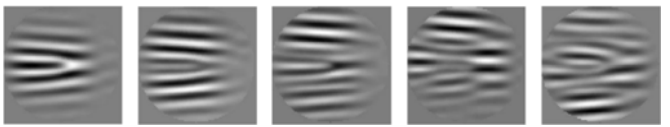

Fig. 7. Five top eigenvectors (from first to fifth rank) of PCA for the texture-based local structures

## 2.4 Feature-level Fusion

Generally, the data fusion in pattern recognition is performed at one of three levels, i.e., score-, decision- and feature-level fusion [36]. Score- and decision-level fusion are more commonly used in biometrics. This is because consolidating information at feature-level is difficult due to the different biometric modalities, which cause inaccessibility and incompatibility [36]. However, both score- and decision-level fusion may fail to reflect original feature characteristics sufficiently, because only the final matching and decision scores are considered in the fusion process. In this paper, the MBLSs and TBLSs are represented in the same subspaces through PCA; thus, feature-level fusion can feasibly consolidate them.

As shown in Fig. 8, the MBLSs and TBLSs are projected by the subspace projection matrix $A_M \in \mathbb{R}^{n_m \times n_p}$ and $A_T \in \mathbb{R}^{n_t \times n_p}$, respectively, where $n_p$ is the reduced dimension. Both reduced-dimension local structure vectors are respectively expressed as $f_{min} = [f_{min_1}, f_{min_2}, \cdots, f_{min_{n_p}}] \in \mathbb{R}^{n_p}$ and $f_{text} = [f_{text_1}, f_{text_2}, \cdots, f_{text_{n_p}}] \in \mathbb{R}^{n_p}$.

Then, each vector is normalized to have the same mean and standard deviation, such that

$$f'_{min_i} = \frac{f_{min_i} - \mu_{min}}{\sigma_{min}},$$
$$f'_{text_i} = \frac{f_{text_i} - \mu_{text}}{\sigma_{text}}, \quad (6)$$

where $\mu_{min}, \sigma_{min}, \mu_{text}$ and $\sigma_{text}$ denote the mean and standard deviation values of the minutia-based and texture-based vectors ($f_{min}, f_{text}$), respectively. The $\omega_M$ and $\omega_T$ weights reflect the relative importance between the minutia- and texture-based features, and were obtained empirically in this work. Thus, the final representation of each minutia is $f_{fused} = [\omega_M(f'_{min_1}, \cdots, f'_{min_{n_p}}), \omega_T(f'_{text_1}, \cdots, f'_{text_{n_p}})] \in \mathbb{R}^{2n_p}$, and an entire fingerprint image is expressed as $\mathcal{F} = [f^1_{fused}, f^2_{fused}, \cdots, f^n_{fused}] \in \mathbb{R}^{n \times 2n_p}$, where $n$ is the number of extracted minutia.

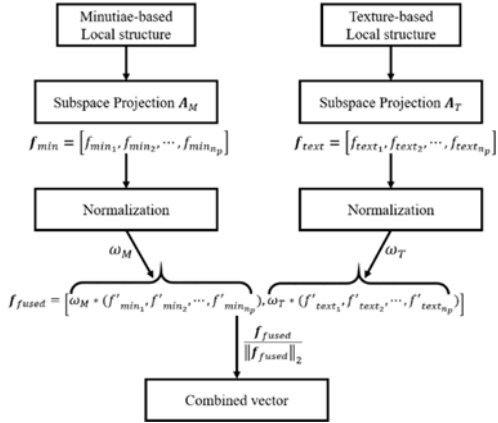

Fig. 8. Feature-level fusion of minutia-based and texture-based local structures

## 2.5 Clustering and bit-string conversion

Note that the size of the intermediate fingerprint representation $\mathcal{F} \in \mathbb{R}^{n \times 2n_p}$, i.e., $n$, remains unfixed, as the minutia number varies according to the fingerprint image. In this section, the mechanism for conversion from $\mathcal{F}$ to a fixed-size bit-string via clustering is described.

We first establish a training pool of fused vectors $f_{fused} \in \mathbb{R}^{2n_p}$, derived from minutiae in the training fingerprint image set. K-means clustering is employed to group similar structures into $K$ clusters $\mathcal{C} = \{C_1, C_2, \cdots, C_K\}$. Each cluster is represented by its corresponding prototype vectors, $c_d \in \mathbb{R}^{2n_p}$, with $d = 1,\ldots,K$, as illustrated in Fig. 9. Although other clustering methods [37] are available, K-means clustering with ED metric is preferred because of its simplicity.

To represent a fingerprint image by a $K$-bit ordered string, $b = [b_1, \ldots, b_d, \ldots, b_K]$, the $d$-th bit in $b$ is set to 1 if an $f_{fused}$ is captured by the $d$-th cluster. In other words, the cluster $d$ to which $f^i_{fused}$ (the fused features of minutia $m_i$) belongs can be determined from

$$d^* = \arg\min_d \|f^i_{fused} - c_d\|_2 \in [1, K], 1 \leq d \leq K, \quad (7)$$

where $d^*$ indicates the $d^*$th bit in $b$.

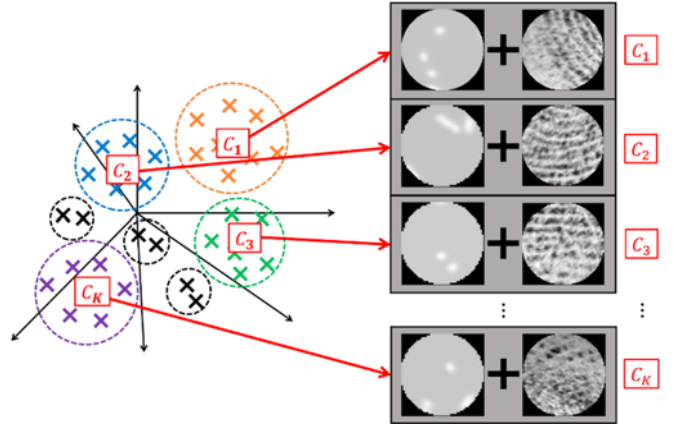

Fig. 9. Clustering of fused vectors by K-means clustering

Generally, each cluster, which manifests as a hypersphere with radius $\mathcal{M}_d$, $d=1,\ldots,K$, has a different size, depending on the $f_{fused}$ embraced by the cluster. Note that the distance of $f_{fused}$ from $c_d$ may not be the shortest for a specific cluster; however, $f_{fused}$ may be included if it is close to the boundary of that cluster. Therefore, in this work, it is instructive to consider both the distance from $c_d$ as well as the cluster size $\mathcal{M}_d$ when determining the most similar cluster to $m_i$ as depicted in Fig. 10.

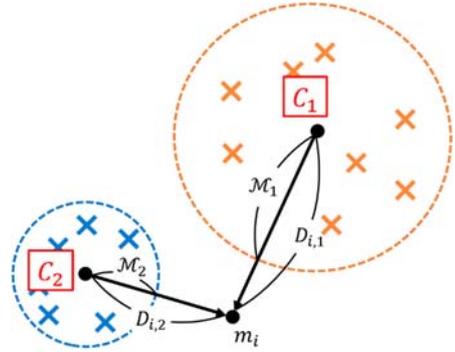

Fig. 10. Distance comparison between clusters ($C_1$ and $C_2$) and minutia $m_i$. Although $m_i$ is closer to $C_2$ ($D_{i,2} < D_{i,1}$), it is more likely to be included in $C_1$ because of the shorter distance from the boundary of the cluster $C_1$ ($D_{i,2} - \mathcal{M}_2 > D_{i,1} - \mathcal{M}_1$)

The value of $\mathcal{M}_d$ can be estimated from the mean value of the EDs between the embraced $f_{fused}$ and $c_d$; however, the mean value may be biased for a small number of $f_{fused}$. Thus, the opposite approach is adopted in our work; that is, $\mathcal{M}_d$ is estimated from the mean ED value between $c_d$ and the $f_{fused}$ located near the boundary of the cluster *but excluded* from the cluster. This approach is deemed more reliable because of the large training instances. Accordingly, $\mathcal{M}_d$ can be estimated as follows:

$$D_{i,d} = \|f^d_{fused,i} - c_d\|_2, 1 \le i \le N_t,$$
$$\{s_{d,1}, s_{d,2}, \cdots, s_{d,N_t}\} \leftarrow sort_{ascending}(D_{1,d}, D_{2,d}, \cdots, D_{i,N_t}),$$
$$\mathcal{M}_d = \frac{1}{N_c} \sum_{i=1}^{N_c} s_{d,i}, \quad (8)$$

where $f^d_{fused,i}$ indicates the $f_{fused}$ located outside the $d$th cluster and $N_t$ is the number of $f^d_{fused,i}$ in the training pool. When $(D_{1,d}, D_{2,d}, \cdots, D_{i,N_t})$ are sorted in ascending order, the mean value of the top $N_c$ ($<N_t$) distances $s_{d,i}$ is considered to be $\mathcal{M}_d$. Therefore, (7) can be reformulated to

$$d^* = \arg\min_d(\|f^i_{fused} - c_d\|_2 - \mathcal{M}_d) \in [1, K], 1 \le d \le K. \quad (9)$$

From (9), we can find the most similar cluster to $f^i_{fused}$ considering both the distance from the center ($\|f^i_{fused} - c_d\|_2$) and $\mathcal{M}_d$. Then, a "1" or "0" value is assigned at the $d^*$ bit position according to the rule

$$b_{d^*} = \begin{cases} 1 & \|f^i_{fused} - c_{d^*}\|_2 - \mathcal{M}_{d^*} < \tau_s, \\ 0 & otherwise \end{cases} \quad (10)$$

where $\tau_s$ is the maximum permitted distance of $f^i_{fused}$ from the boundary of the selected $d^*$th cluster if the "1" value is finally set to the $d^*$th bit in $b$. Equation (10) is designed to eliminate outlier instances where the $f_{fused}$ of interest and the cluster are excessively separated, even though the cluster is selected as the most similar among $K$ clusters. The bit assignment procedure is repeated for all $f_{fused}$ derived from a fingerprint image. Hence, the input fingerprint image is eventually represented as an ordered bit-string with length $K$, $b=[b_1, \ldots, b_d, \ldots, b_K]$.

### 2.6 Finger-specific Bit-training

In this subsection, we outline a finger-specific bit-training algorithm, which is intended to further improve the accuracy. The proposed algorithm requires fingerprint samples of same finger to select useful bit positions, which will be used for fingerprint matching process. This method is inspired by reliability-based dynamic quantization [38]. However, our bit selection is determined by two criteria, discrimination power and reliability, which can capture the inter- and intra-class variances of the fingerprint bit-string, respectively.

#### 2.6.1 Discrimination Power Criteria

In this work, the discrimination power of a cluster is determined by considering both the global discrimination power (GDP) and inter-class variance. The cluster GDP can be estimated by measuring the cluster *cardinality*, which is the number of $f_{fused}$ belonging to that cluster. Low cluster cardinality implies that the cluster is more discriminative, because only a small number of $f_{fused}$ possess similar characteristics to the cluster. Formally, the cardinality of the $d$th cluster $H_d$ is defined as:

$$H_d = card\left(\arg\min_t \left(\|f^i_{fused} - c_t\|_2 - \mathcal{M}_t\right) = d\right), 1 \le t \le K \quad (11)$$
$$\mathbf{H} = [H_1, \ldots, H_d, \cdots, H_K],$$

where card() denotes the cardinality of each cluster. Here, a small $H_d$ suggests high GDP, and vice versa. Neverthe-less, $H_d$ cannot be used directly, because a change of training sample may cause $H_d$ variation. To resolve this problem, $H_d$ is normalized as follows:

$$\widehat{H}_d = 1 - \frac{H_d - \min(\mathbf{H})}{\max(\mathbf{H}) - \min(\mathbf{H})}. \quad (12)$$

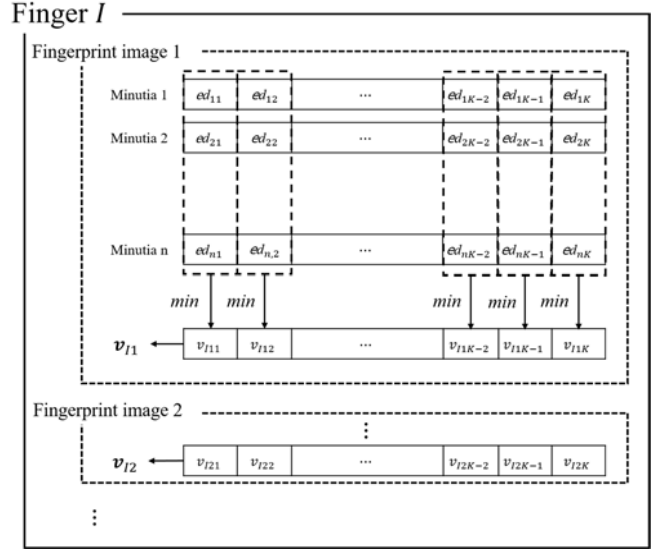

Fig. 11. Derivation of fixed length vector $v_{Ij}$ from $k$th fingerprint image of finger $I$. Here, $ed_{ij}$ indicates the Euclidean distance between $f^i_{fused}$ and the $j$th cluster.

Accordingly, the $\widehat{H}_d$ of the $d$th cluster is in the range of [0, 1], and "1" indicates the maximum GDP.

In addition to the GDP, the inter-class variance of a cluster can also be exploited to measure the discriminative power of the fingerprint bit-string. Prior to bit-string conversion via (9) and (10), each $\mathcal{F}$ is represented as a real-valued vector composed of EDs, which can be utilized for feature values. Suppose $v_{Ijd}$ is the minimum ED between the $d$th $c_d$ and all fused vectors $f_{fused}$ obtained from the $j$th fingerprint image of finger $I$, where $1 \le d \le K$. Then, a fixed-length vector $v_{Ij} = [v_{Ij1}, v_{Ij2}, \cdots, v_{IjK}] \in \mathbb{R}^K$ can be derived (see Fig. 11). Small $v_{Ijd}$ suggests that finger $I$ has a more distinct feature for the corresponding cluster; hence the ED vector $v_{Ij}$ can be used to represent the minutia features of fingerprint image $j$.

Let $N_I$ denote the number of training fingerprint image samples of finger $I$; then, the mean feature vector of $I$, $\mu_I$, is given as

$$\mu_{Id} = \frac{1}{N_I} \sum_{j=1}^{N_I} v_{Ijd}, \quad (13)$$
$$\mu_I = [\mu_{I1}, \mu_{I2}, \cdots, \mu_{IK}].$$

Next, a global mean vector $\mu^G$ can be computed from

$$\mu^G_d = \frac{1}{N_t} \sum_{I=1}^{N_t} \mu_{Id}, \quad (14)$$
$$\mu^G = [\mu^G_1, \mu^G_2, \cdots, \mu^G_K],$$

where $N_t$ is the total number of fingers in the training set. As the global mean vector in (14) is pre-computed from training samples and saved with clusters, it can be utilized to select useful bits for a given fingerprint image.

Accordingly, the inter-class variance of each bit in $I$ is given as follows:

$$x_{Ijd} = \begin{cases} v_{Ijd} - \mu_d^G & (v_{Ijd} - \mu_d^G) < 0 \\ 0 & otherwise \end{cases},$$
$$v_{Id}^{BV} = \frac{1}{N_I}\sum_{j=1}^{N_I}(x_{Ijd})^2, \qquad (15)$$
$$\boldsymbol{v}_I^{BV} = [v_{I1}^{BV}, v_{I2}^{BV}, \cdots, v_{IK}^{BV}].$$

In (15), $x_{Ijd}$ is employed so that only the $v_{Ijd}$ with values smaller than $\mu_d^G$ are used; this is deemed discriminative for computation of the inter-class variance $v_{Id}^{BV}$. Accordingly, high $v_{Id}^{BV}$ implies that cluster $d$ of finger $I$ has more distinct features than other fingers. Finally, the discrimination power vector of each bit of $I$ is obtained by combining the GDP (12) and the inter-class variance (15), as follows:

$$\boldsymbol{p}_I = [p_{I1}, p_{I2}, \cdots, p_{IK}] = [\hat{H}_1 v_{I1}^{BV}, \hat{H}_2 v_{I2}^{BV}, \cdots, \hat{H}_K v_{IK}^{BV}]. \quad (16)$$

### 2.6.2 Reliability Criterion

The reliability criterion reflects the consistency of the binary value in each bit position for multiple fingerprint images, to reduce false acceptance of impostor bits assigned during the bit-conversion process. Let $b_{Ijd}$ denote the $d$th bit of the $j$th fingerprint image from the $I$th finger; then, the reliability vector $\boldsymbol{l}_I$ is defined as

$$\boldsymbol{l}_I = [l_{I1}, l_{I2}, \cdots, l_{IK}] = \left[\frac{\sum_{j=1}^{N_I} b_{Ij1}}{N_I}, \frac{\sum_{j=1}^{N_I} b_{Ij2}}{N_I}, \cdots, \frac{\sum_{j=1}^{N_I} b_{IjK}}{N_I}\right]. \quad (17)$$

### 2.6.3 Bit selection based on discrimination power and reliability

After the bit-string discrimination power is determined from (16), a bit mask having the same size as the bit-string is derived to access the bit-string reliability (17). When the bit positions of $I$ are sorted by their discrimination power values ($\boldsymbol{p}_I$) in descending order, starting from the first, if the reliability of the $t$th bit is larger than an adaptive threshold value $\tau_{It}$, this bit is set to 1 and participates in matching; otherwise, it is set to 0 and masked. This process is repeated for the remaining bits in the bit-string. The value of $\tau_{It}$ is computed adaptively with respect to $t$ as follows:

$$\tau_{It} = \alpha + \frac{(1-\alpha)}{1 + e^{-\beta(t - N_{mean})}}, \quad (18)$$

where $\alpha$ and $\beta$ are the empirical parameters and $N_{mean}$ is the average minutia number extracted from multiple samples of $I$. Thus, according to (18), the closer the bit-training process iterates to $N_{mean}$, the harder the selection of a new bit for participation in matching becomes. Hence, only reliable and discriminative bits are selected and participate in the bit-string matching process, further improving the accuracy.

The overall process of proposed fingerprint-specific bit-training is summarized in Algorithm 1.

## 2.7 Fingerprint Matching before and after Bit-string Conversion

Although bit-string conversion of a fingerprint image is the main purpose of our work, the recognition accuracy of the fused local structure (Section 2.4) is also assessed. This evaluation is beneficial to examine the performance gain due to the fusion of the MBLSs and TBLSs, and the performance change due to bit-string conversion. Therefore, in this section, we present two matching methods used before and after the bit-string conversion.

---

**Algorithm 1** Pseudo code of fingerprint-specific bit-training

◆ **Input**:
$\boldsymbol{v}_I$: Euclidean distance feature vectors of finger $I$
$\boldsymbol{b}_I$: Ordered and fixed-length bit-strings of finger $I$
$K$: Cluster number

◆ **Initialize**:
$\boldsymbol{m}_{sel}(1:K) = [0]$

◆ **Do**:
$\boldsymbol{p}_I \leftarrow (\boldsymbol{v}_I)$ % $\boldsymbol{p}_I$: Discrimination power vector of finger $I$ (16).
$\boldsymbol{l}_I \leftarrow (\boldsymbol{b}_I)$ % $\boldsymbol{l}_I$: Reliability vector of finger $I$ (17)
$\boldsymbol{p}'_I = sort_{descending}(\boldsymbol{p}_I)$
For $t \leftarrow 1:K$
  $idx = $ bit_position($\boldsymbol{p}'_I(t)$)
  If $\boldsymbol{l}_I(idx) > \tau_{It}$   % $\tau_{It} = \alpha + \frac{(1-\alpha)}{1+e^{-\beta(t-N_{mean})}}$
    $\boldsymbol{m}_{sel}(idx) = 1$
  End
End

◆ **Output**: $\boldsymbol{m}_{sel}$ ($K$-dimensional binary bit-string for use in fingerprint matching)

---

### 2.7.1 Fingerprint matching in PCA subspace before bit-string conversion

The dissimilarity measure of two PCA-reduced dimension fused vectors described in Section 2.4 is measured by the ED. Then, the local greedy similarity (LGS) method [39], which measures the average dissimilarity score (ED in this context) of $n_L$ fused vector pairs starting from the highest similarity, is adopted to find the final matching score. Here, $n_L$ can be computed as follows:

$$n_L = min_{n_L} + \left\lfloor \frac{max_{n_L} - min_{n_L}}{1 + e^{-\tau_P(\min\{N_A, N_B\} - \mu_P)}} \right\rfloor \quad (19)$$

where $N_A$ and $N_B$ are the numbers of fused vectors extracted from fingerprints $I_A$ and $I_B$, respectively, and $\tau_P, \mu_P, max_{n_L}$ and $min_{n_L}$ are empirical parameters. Then, the matching score of fingerprints $I_A$ and $I_B$ is computed as follows:

$$S(I_A, I_B) = \frac{\sum_{i=1}^{n_L} \mathcal{D}_i^{asc}(\mathcal{F}^A, \mathcal{F}^B)}{n_L} \quad (20)$$

where $\mathcal{F}^A$ and $\mathcal{F}^B$ are fused vectors obtained from $I_A$ and $I_B$, and $\mathcal{D}_i^{asc}$ denotes the ED between the $i$th fused vector pair when all fused vector pairs of $I_A$ and $I_B$ are sorted by the dissimilarity score in ascending order.

### 2.7.2 Fingerprint matching after bit-string conversion

The normalized intersection score [25] is used for fingerprint matching after bit-string conversion, measuring the intersection of two bit-strings normalized by the total number of "1"s in the bit-strings. The score is given as follows:

$$S_b(\boldsymbol{b}^A, \boldsymbol{b}^B) = \frac{(n^A + n^B)\sum_{i=1}^{K}(b_i^A \text{ AND } b_i^B)}{(n^A)^2 + (n^B)^2},$$
$$n^A = \sum_{i=1}^{K} b_i^A, n^B = \sum_{i=1}^{K} b_i^B, \quad (21)$$

where $\boldsymbol{b}^A$ and $\boldsymbol{b}^B$ are the bit-strings of fingerprint images $A$ and $B$ respectively; $b_i^A, b_i^B$ indicate the $i$th bit elements of $\boldsymbol{b}^A$ and $\boldsymbol{b}^B$, respectively; and $\sum_{i=1}^{K}(b_i^A \text{ AND } b_i^B)$ is the total number of matched bits for which both $\boldsymbol{b}^A$ and $\boldsymbol{b}^B$ have "1" values at the same bit position. Note that $S_b(\boldsymbol{b}^A, \boldsymbol{b}^B)$ in (21) ranges from 0 to 1, and a "1" score indicates a perfect match of two bit-strings.

## 3 EXPERIMENTAL RESULTS

### 3.1 Databases and Parameters

The performance of the proposed method (PM) was evaluated on seven databases of the Fingerprint Verification Competition (FVC), i.e., FVC2002 (DB1, DB2 and DB3), FVC 2004 (DB1 and DB2), and FVC2006 (DB2 and DB3). The FVC2002 and FVC2004 datasets are composed of 100 subjects with eight samples per subject (total: 800 images), and the FVC2006 datasets have 140 subjects with 12 samples per subject (total: 1,680 images). *VeriFinger 6.2* [40] was used to locate minutiae in the fingerprint images, and the extracted minutia templates followed the ISO/IEC 19794-2 [41] format. For performance evaluation of the PM, the equal error rate (EER) and receiver operating characteristic (ROC) were observed.

The parameters used in the PM are listed in Table 2. The optimal parameters for local structure and feature-level fusion were obtained experimentally with the images from Set_B of FVC datasets (FVC2002, FVC2004 and FVC2006). $n_m$ is determined by 10 times downscaling the MBLS area i.e. $n_m \approx \pi r_m^2/10$. The MBLS radius ($r_m$ = 80 pixels) and the TBLS radius ($r_m$ = 40 pixels) were determined when the system achieves the best accuracy. The optimal weight for the MBLS in feature-level fusion $\omega_M$ was set to 0.6, which is larger than $\omega_T$. This is due to the minutia are deemed more robust and discriminative than the texture information. The number of clusters $K$ varied with respect to different datasets, but $K$ is initially set as 10% of the total number of fused vectors obtained from the training dataset and then is fine-tuned with reference to the best training accuracy performance. Thus, the number of clusters is training datasets dependence. The other parameter values ($\tau_s, \alpha, \beta, max_{n_P}, min_{n_P}, \mu_P, \tau_P$) were also determined based on the Set_B of FVC datasets.

### 3.2 Performance Analysis

The PM performance was evaluated for three cases: 1) before bit-string conversion, 2) after ordered and fixed-length bit-string conversion, and 3) after application of the finger-specific bit-training algorithm to the bit-strings. The first case indicated the accuracy of the fused MBLSs and TBLSs. The accuracy of the fixed-length bit-strings was compared with those of previous methods in the second case. In the last case, the accuracy after bit-training was examined.

#### 3.2.1 Performance before bit-string conversion

Genuine and impostor matching distributions were observed according to the original FVC protocol [35]. As suggested by the protocol, each image of a subject was compared to the remaining images of the same subject for genuine matching. The first image of each subject was compared with the first images of the remaining subjects for impostor matching. Therefore, 2,800 ($\frac{8\times7}{2} \times 100$) genuine and 4,950 ($\frac{100\times99}{2}$) impostor matching scores were obtained from FVC2002 and FVC2004, respectively, with 9,240 ($\frac{12\times11}{2} \times 140$) genuine and 9,730 ($\frac{140\times139}{2}$) impostor matching scores from FVC2006.

The experimental results are presented in Tables 3 and 4, and Fig. 12. The complex Gaussian mixture model proposed by Liu et al. [42] was considered, which expresses the positions of the minutia based on the probability distribution model. In addition, the MCC was employed, for which the parameter values given by MCC SDK 2.0 [43] were adopted in this test. In the tables, the results obtained for the PM using the MBLS alone (Section 2.2) and those obtained from the fusion of the MBLSs and TBLSs (Section 2.4) are marked as the "PM (minutiae only)" and "PM (combined)," respectively.

From Tables 3 and 4, the following results were deter-

TABLE 2
EMPIRICAL PARAMETER VALUES SET IN EXPERIMENTS

| Phase | Parameters | Description | Value |
|---|---|---|---|
| Local structure | $r_m$ | Minutia-based local structure radius (in pixel) | 80 |
| | $n_m$ | Dimensional size of minutia-based local structure | 2,006 |
| | $r_t$ | Texture-based local structure radius (in pixel) | 40 |
| | $n_t$ | Dimensional size of texture-based local structure | 5023 |
| | $n_p$ | PCA dimension size | 100 |
| | | PCA dimension size only for Case 1 | 50 |
| Bit-conversion | $\tau_s$ | Parameter in (10) for Case 1 | -0.05 |
| | | Parameter in (10) for Case 2 | -0.11 |
| Bit-training | $\alpha, \beta$ | Parameters in (18) | 0.45, 0.4 |
| Matching | $max_{n_L}, min_{n_L}$ | Parameters in (19) | 10, 4 |
| | $\mu_P, \tau_P$ | Parameters in (19) | 35, 0.4 |
| Feature fusion | $\omega_M, \omega_T$ | Weights for feature-level fusion | 0.6, 0.4 |
| Clustering | K | Number of clusters for Case 1 | 15,000 |
| | | Number of clusters for Case 2 | 4,500 |
| | $N_c$ | Parameter in (8) | 300 |

TABLE 3
COMPARISONS WITH PM BEFORE BIT-STRING CONVERSION ON FVC2002 (EQUAL ERROR RATE (%))

| Method | FVC2002 | | |
|---|---|---|---|
| | DB1 | DB2 | DB3 |
| CGMM [42] | 2.21 | - | - |
| MCC [20] | 1.25 | 0.89 | 3.28 |
| PM (minutiae only) | 1.49 | 1.27 | 5.59 |
| PM (combined) | **1.10** | **0.70** | **4.23** |

TABLE 4
COMPARISONS WITH PM BEFORE BIT-STRING CONVERSION ON FVC2004 AND FVC2006 (EQUAL ERROR RATE (%))

| Methods | FVC2004 | | FVC2006 | |
|---|---|---|---|---|
| | DB1 | DB2 | DB2 | DB3 |
| CGMM [42] | 8.92 | 9.56 | - | - |
| MCC [20] | 8.24 | 7.32 | 1.84 | 8.13 |
| PM (minutiae only) | 8.76 | 6.92 | 1.55 | 8.19 |
| PM (combined) | **7.71** | **5.79** | **0.82** | **5.24** |

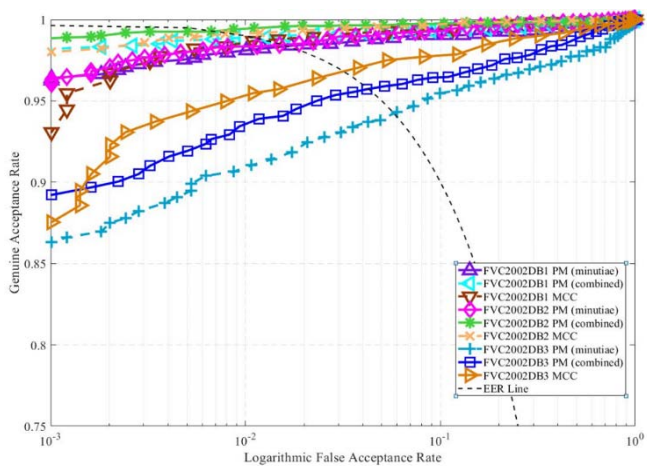

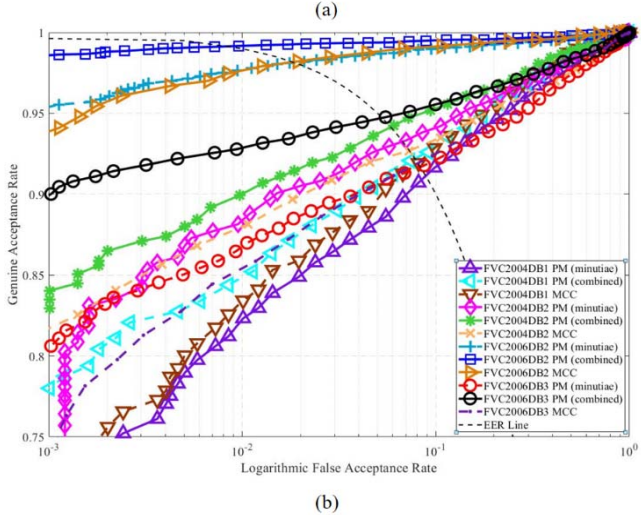

Fig. 12. ROC curves of PM before bit-string conversion for (a) FVC2002 and (b) FVC2004 and FVC 2006 DBs

mined:
1. PM (minutiae only) exhibited superior performance to CGMM, which also uses local minutia information only. This superiority is due to the robustness of the proposed 2DGF modeling to local errors such as alignment errors and skin distortion (Section 2.2).
2. The PM combining the MBLSs and TBLSs exhibited the lowest EER for all datasets except FVC2002 DB3. This result shows that the MBLS utilizing the elliptical Gaussian function to be robust to minutia positioning error successfully represent neighbor minutia positions and combination with the TBLS has good synergy to extract the discriminative minutia feature.
3. For FVC2002 DB3, the fingerprint images captured from small parts of fingerprints containing limited numbers of minutia. Thus, due to the lack of minutia information, the MCC (utilizing directional information of neighbor minutiae) achieved better performance than the PM.
4. PM (combined) yielded superior performance compared to PM (minutiae only). This shows that even incorporation of simple texture information can improve the performance notably.

*3.2.2 Performance analysis after bit-string conversion*

For comparison with the existing methods, two different cases were considered. In the first case, to examine the generalization capability of the PM, clusters were created from training datasets, and the test samples were chosen from different datasets. For the second case, as in [24], images captured from a finger were used for cluster creation and other images from the same finger were used for testing. This was for comparison with the method reported in [4] on the basis of similar conditions.

3.2.2.1 Case 1: Clustering and testing on different datasets
In this case, as in the conventional performance test, the datasets for clusters and tests were different. In addition to FVC datasets, we generated random minutiae-based local structures to augment the clusters pool. Thus, the generated clusters can cover the extended minutia feature area, which may not be reached by FVC datasets. In addition, Set_B fingerprint images of FVC datasets (FVC2002, 2004 and 2006) are also included as training data.

TABLE 5
COMPARISONS OF PMs ON FVC2002 DB1 AFTER BIT-STRING CONVERSION (DIFFERENT DATASETS ARE USED FOR CLUSTERING AND TESTING) (EQUAL ERROR RATE (%))

| Method | Training data for clustering | Number of bits | Accuracy (EER) |
|---|---|---|---|
| Pair-minutiae [13], [14], [19] | N/A | 32,768 | 15.5 |
| PM (minutiae) | FVC2002 (DB2, DB3) + FVC2004 (DB1, DB2) + FVC2006 (DB2, DB3) | 15,000 | 7.1 |
| PM (combined) | | | 6.67 |
| PM (bit-training) | FVC Set_B + Random minutiae | | **3.95** |

TABLE 6
COMPARISONS OF PMs ON FVC2002 DB2 AFTER BIT-STRING CONVERSION (DIFFERENT DATASETS ARE USED FOR CLUSTERING AND TESTING) (EQUAL ERROR RATE (%))

| Method | Training data for clustering | Number of bits | Accuracy (EER) |
|---|---|---|---|
| Bringer [31] | N/A | 50,000 | 5.3 |
| Pair-minutiae [13], [14], [19] | N/A | 32,768 | 17.6 |
| PM (minutiae) | FVC2002 (DB1, DB3) + FVC2004 (DB1, DB2) + FVC2006 (DB2, DB3) + FVC Set_B + Random minutiae | 15,000 | 4.38 |
| PM (combined) | | | 4.05 |
| PM (bit-training) | | | **2.46** |

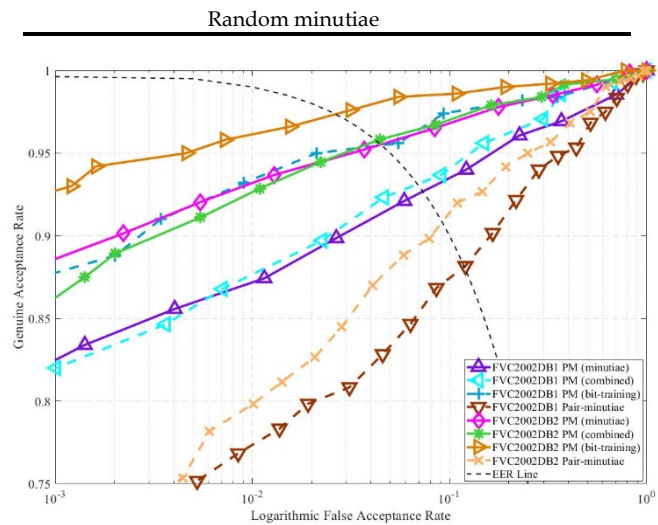

Fig. 13. ROC curves of PM after bit-string conversion

The chosen PCA dimension for both MBLSs and TBLSs is set to 50, whereby was determined experimentally. Compared to other evaluations, the PCA dimension for Case 1 is reduced for computational efficiency, because numerous clusters (15,000) are needed to represent various minutia characteristics. The feature-level fusion was processed with the same weights and feature normalization introduced in Section 2.4. For each input minutia, five clusters from top similarity are selected and '1' values are assigned at those positions using Eq. (9). Genuine and impostor matching scores were obtained according to the original FVC protocol, except for the bit-training experiment. For bit-training, three fingerprint images (samples 1–3) of each finger were used as training samples to obtain the bit mask, and other fingerprint images (samples 4–8) were used as test samples.

In Tables 5 and 6, the performance of the proposed bit-string conversion method is compared with existing methods, and Fig. 13 shows the ROC curves of the test results. The pair-minutiae approach [13], [14], [19] was considered, with the same parameter values reported in the cited studies being used in our tests. In addition, three different cases of the PM were examined: PM (minutiae only) and PM (combined), as defined above, and the case after implementation of the finger-specific bit-training process (PM (bit-training)).

The following evaluation results were obtained:
1. The PM exhibited superior accuracy to the Pair-minutiae [13], [14], [19] and Bringer [31] methods. The bit-string proposed in our method has 15,000 bit size. However, those of the existing methods were 32,768 [13], [14], [19] and 50,000 [31] bits, respectively. Thus, the result shows that the clustering of similar local structures yields good performance while requiring fewer bits.
2. The performance was improved when the MBLSs and TBLSs were fused. In the PM, fusion of two structures is simple, as both are represented in the same domain, and the results showed that the fused structure reveals more detailed information on the neighboring area of a minutia. However, the pixel intensity used in the TBLS in our work is vulnerable to the alignment error for the center minutia or finger elastic skin distortion. Thus, a more sophisticated feature such as the pixel gradient or ridge shape feature [44] may be considered for greater accuracy improvement.
3. The accuracy was further improved by the finger-specific bit-training. Although this result is unfair because different testing sets were used for the experiments with and without bit-training, the accuracy improvement was quite large. Thus, the result indicates that the discriminative and reliable bits were properly selected by the proposed bit-training method. The performance improvement through bit-training will be especially useful for small mobile fingerprint scanners, where many fingerprint images are input for the enrollment of one finger.
4. By comparing Tables 5 and 6 to Tables 3 and 4, it is noted that the performance degraded slightly when the fingerprint image was converted to the bit-string representation. One of the major reasons for this is that the training datasets or clusters used were not adequate to cover all possible local structures; thus, if clusters were constructed from more training data, the accuracy could be improved.

3.2.2.2 Case 2: Different clustering and testing from the same subject
In this case, to compare the PM performance with existing methods [4], [33], experiments were performed under similar conditions to [4]. That is, three fingerprint images from each finger were used as training samples for the clustering and another five images from the same finger were used for testing. All experiments were performed with 4,500 clusters, and the three training samples were also used for bit-training.
Tables 7 and 8 compare the PM recognition accuracy with previous methods when fingerprint images are represented in ordered and fixed-length bit-strings. Fig. 14 shows the ROC curves of the test results. Here, pair-minutiae

TABLE 7
COMPARISONS OF PMS AFTER BIT-STRING CONVERSION
(EQUAL ERROR RATE (%))

| Methods | FVC2002 | | |
|---|---|---|---|
| | DB1 | DB2 | DB3 |
| Pair-minutiae [13], [14], [19] | 11.8 | 9.08 | 10.84 |
| Xu [33] | - | 3.0 | - |
| Jin [4] | 0.44 | 0.33 | 4.17 |
| PM (combined) | 1.08 | 0.99 | 3.39 |
| PM (bit-training) | **0.39** | **0.33** | **0.73** |

TABLE 8
COMPARISONS OF PMS AFTER BIT-STRING CONVERSION
(EQUAL ERROR RATE (%))

| Methods | FVC2004 | | FVC2006 | |
|---|---|---|---|---|
| | DB1 | DB2 | DB2 | DB3 |
| Pair-minutiae [13], [14], [19] | 14.8 | 14.56 | 4.67 | 12.05 |
| Xu [33] | - | - | - | - |
| Jin [4] | 4.56 | 5.28 | - | - |
| PM (combined) | 9.22 | 7.27 | 1.12 | 7.25 |
| PM (bit-training) | **3.97** | **2.77** | **0.42** | **3.91** |

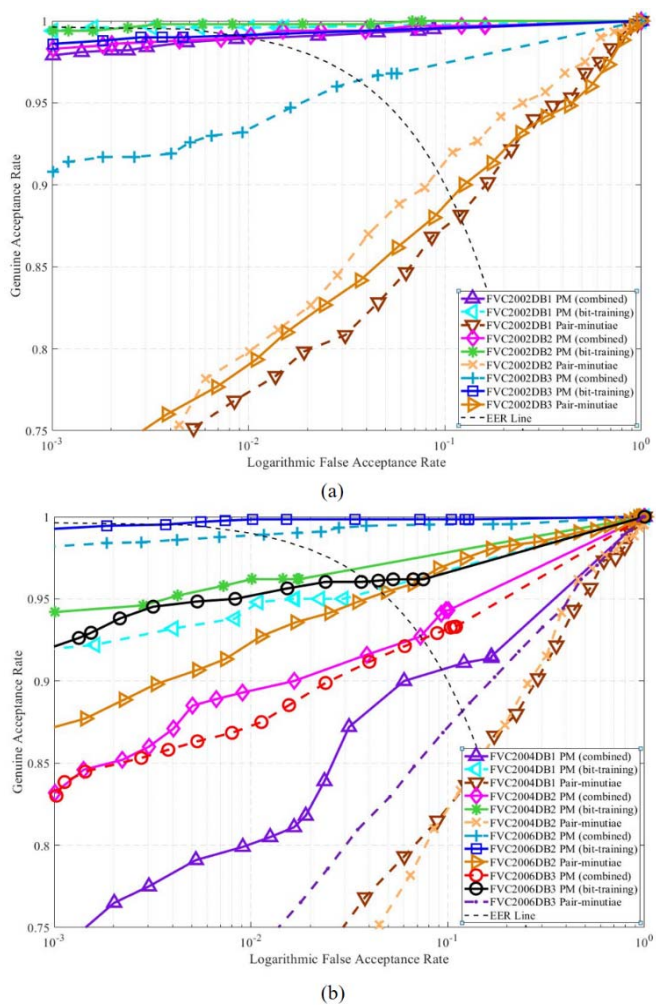

Fig. 14. ROC curves of PM after bit-string conversion matching.

[13], [14], [19] were considered, where the same parameter values reported in the cited studies were used in our tests. The Xu [33] method was also considered. Different from our PM, only four fingerprint image samples from each finger were used for the tests in [33]. Further, the technique reported by Jin et al. [4] was considered, in which multiple fingerprint samples of a finger were used for the training process and a bit-string was generated. Here, PM (combined) is the result of the fixed-length bit-conversion using the combined structures, and PM (bit-training) is the result after finger-specific bit-training is applied to PM (combined).

The following observations were made based on the experimental results:

1. Except for FVC2002 DB2, PM (bit-training) exhibited the lowest EERs. These results show that the proposed fixed-length bit-string conversion method achieves superior recognition accuracy with clustering of similar local structures, where the simple ED measures the dissimilarity among minutiae.
2. As apparent from Tables 7 and 8, the EER decreased notably when finger-specific bit-training was applied. The results show the superiority of the proposed bit-training process, which selects the discriminative and reliable bits for fingerprint matching.
3. The pair-minutiae [13], [14], [19] exhibited the lowest performance over all datasets. This means that the quantization of feature values for fingerprint bit-conversion loses the discriminative information of original features. In contrast, the PM achieved superior performance with fewer bits.

### 3.3 Analysis of Computational Times

As shown in Table 9, only 0.05 seconds for enrollment and 0.0006 seconds for matching are required for the TBLS. This result suggests that the TBLS is reasonable to be combined with the MBLS when we consider its accuracy improvement obtained from all tests. In addition, we

TABLE 9
AVERAGE COMPUTATIONAL TIME AND TEMPLATE SIZE OF THE PROPOSED METHOD BEFORE BIT-STRING CONVERSION (SECONDS)

| Operation | | FVC2002 | | | FVC2004 | | FVC2006 | |
|---|---|---|---|---|---|---|---|---|
| | | DB1 | DB2 | DB3 | DB1 | DB2 | DB2 | DB3 |
| Enrollment | Create MBLS | 0.2290 | 0.2025 | 0.1833 | 0.2278 | 0.2409 | 0.2279 | 0.2824 |
| | Create TBLS | 0.0566 | 0.0562 | 0.0452 | 0.0553 | 0.0591 | 0.0556 | 0.0675 |
| Matching | MBLS | 0.0044 | 0.0032 | 0.0029 | 0.0051 | 0.0049 | 0.0072 | 0.0081 |
| | Combined | 0.0059 | 0.0038 | 0.0035 | 0.0060 | 0.0059 | 0.0079 | 0.0087 |
| Template size (Bytes) | | 25,600 | 20,800 | 18,400 | 28,800 | 24,000 | 31,200 | 30,400 |

414TABLE 10
COMPARISON OF AVERAGE COMPUTATIONAL TIMES OF FINGER-
PRINT BIT-CONVERSION METHODS

| Methods | Average time (in seconds) | | Bit-size |
|---|---|---|---|
| | Generation of bit-string | Matching by bit-strings | |
| Ferrara [20] | 0.017 | 3.0e-03 | 61,040 bits (Variable size) (Average in FVC2006 DB2) |
| Pair-minutiae [13], [14], [19] | 0.04 | 0.5e-03 | 32,768 bits |
| Bringer [31] | N/A | N/A | 50,000 bits |
| Jin [5] | 4 | 1.0e-06 | 200~300 bits |
| PM | 0.6 | 1.0e-04 | 15,000 bits (case 1, Section 3.2.2.1) 4,500 bits (case 2, Section 3.2.2.2) |

can notice that the proposed method is feasible for real-time applications before bit-string conversion. However, the template size is considerably big, so the fixed-length bit-conversion is needed for the storage compression.

In Table 10, the processing times for the two phases in the PM are compared with those of other methods. The times were measured for the following computer specifications: Intel i7 CPU, 8 GB RAM, and MATLAB 2017.

As shown in Table 10, even though MCC takes much smaller time for bit-string conversion than our methods, the variable size bit-strings generated from the MCC requires longer matching process and bigger storage capacity. Therefore, if massive fingerprint data is saved and compared to input fingerprint image, the proposed method has more advantages than the MCC.

Although the method of [4] generates the smallest bit size, the average time for bit conversion is much longer (2–6 s) than the PM (average 0.4–0.8 s). In addition, in the PM, the number of clusters is 15,000, being smaller than those for the method of [31]. In addition, the similarity scores between the minutiae and clusters were computed by a complex exponential computation in the [31] method, but the PM used the simple ED. Therefore, the PM is likely more suited to real-time applications with better recognition accuracy.

### 3.4 Bit-string Compression

As apparent from Table 10, the PM requires 4,500–15,000 bits to represent a fingerprint image according to the cluster numbers set in the K-means clustering algorithm. Although the bit size is significantly lower than most existing methods [13], [14], [31], [19], it remains large in the context of practical usage. However, the bit-string generated by the PM is highly sparse because of the small number of minutiae in a fingerprint image. For instance, the maximum numbers of non-zero bits in any bit-string in FVC2002 DB2 and FVC2002 DB1 are at most 335 and 200, respectively, corresponding to only 2.80% and 1.65% of the bit size.

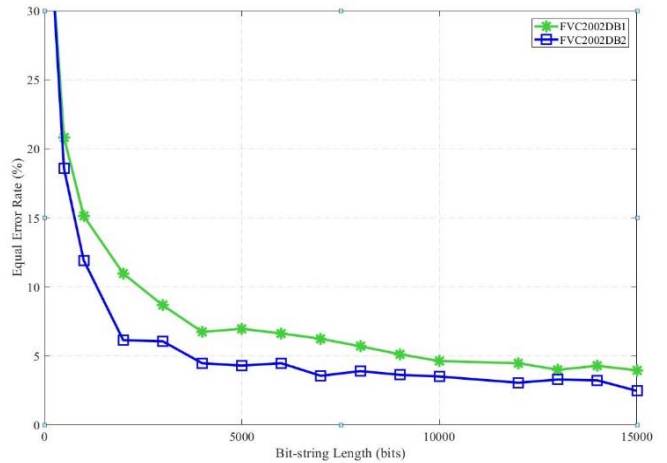

Fig. 15. Accuracy for varying compression length

According to [45], [46], bit-string sparsity is beneficial for compression without accuracy performance degradation. In this study, we employed the binary compression scheme (BCS) introduced in [45], [46] for bit-string compression. Fig. 15 shows the results for varying bit size in the range of 100–15000 bits taken from case 1 experiments (Section 3.2.2.1). For both FVC2002 DB1 and DB2, the accuracy was not affected significantly until the bit-string was compressed to 4,500–5,000 bits, which is a reasonable size in practice.

## 4 CONCLUSION

In this paper, an ordered and fixed-length bit-string conversion method is proposed to address the limitations of conventional fingerprint recognition systems based on minutia representation. A novel minutia-based local structure realized by the normalized mixture of the Gaussian function model is devised, so that the similarities among adjacent minutiae could be measured in terms of the Euclidean distance, and PCA is adopted to reduce the feature dimensions. In addition, the texture information of fingerprint images is expressed by the texture-based local structures, which are also projected into the PCA subspace in the proposed approach. The two types of local structure are fused at the feature level, and K-means clustering is adopted to group similar local structures into clusters. A finger-specific bit-training method is also implemented to improve the overall recognition accuracy by selecting reliable and discriminative bits. Validation experiments were conducted, with the results showing that the proposed method yields superior recognition accuracy than existing methods.

Nevertheless, the proposed method has a small number of limitations, which can be addressed in future work. First, additional minutia information such as data on the type and orientation can be utilized. Generally, consideration of both minutia type and orientation can facilitate improved recognition accuracy when combined with minutia positional information. Second, in this study, the simple pixel intensity was used for the texture-based local structure. However, this representation is prone to image misalignment. Therefore, a more sophisticated feature such as the pixel gradient or the ridge shape feature could be considered. Third, K-means clustering was used to group similar minutia features, but more advanced clustering methods may produce improved and/or more clusters to cover all possible local structures. Finally, matching methods for bit-strings different to that used in our method may improve the accuracy of this technique.

## ACKNOWLEDGMENT

This research was supported by Multi-Ministry Collaborative R&D Program(R&D program for complex cognitive technology) through the National Research Foundation of Korea(NRF) funded by MSIT, MOTIE, KNPA(NRF-2018M3E3A1057289)

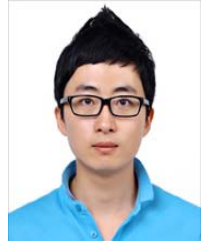

**Jun Beom Kho** received the B.S degree in electrical and electronic engineering from Yonsei University, Seoul, Korea, in 2012. He is currently working toward the M.S.-Ph.D. joint degree in electrical and electronic engineering from Yonsei University, Seoul, Korea. His research interests include biometrics, image processing, and deep learning.

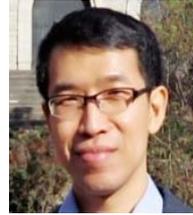

**Andrew Beng Jin Teoh** (SM'12) obtained his BEng (Electronic) in 1999 and Ph.D. degree in 2003 from National University of Malaysia. He is currently an associate professor in Electrical and Electronic Engineering Department, College Engineering of Yon-sei University, South Korea. His research interests are Biometrcs and Machine Learning. He has published more than 280 international refereed journal papers, conference articles, edited several book chapters and edited book volumes. He served and is serving as a guest editor of IEEE Signal Processing Magazine, associate editor of IEEE Transaction of Information Forensic and Security, IEEE Biometrics Compendium and editor-in- chief of IEEE Biometrics Council Newsletter.

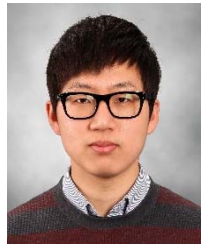

**Wonjune Lee** received the B.S. degree in electrical and electronics engineering in 2009, and the M.S.-Ph.D joint degree in 2017, from Yonsei University, Seoul, South Korea. Currently, he is a research engineer in Hyundai Mobis, Yong-In, South Korea. His research interests include biometrics, computer vision, image processing, and pattern recognition.

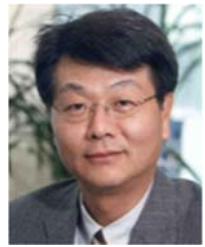

**Jaihie Kim** received Ph.D. from Case Western Reserve University, USA. Since 1984, he has been a professor at the School of Electrical and Electronic Engineering at Yonsei University. He was the Director of Biometric Engineering Research Center in Korea and his research interests include mobile biometrics, spoof detection, cancellable biometrics, and face age estimation/synthesis. More information on Prof. Kim can be found in http://cherup.yonsei.ac.kr/.